\documentclass[sn-basic]{sn-jnl}
\usepackage{graphicx}%
\usepackage{hhline, multirow}%
\usepackage{amsmath,amssymb,amsfonts}%
\usepackage{mathrsfs}%
\usepackage[title]{appendix}%
\usepackage{xcolor}%
\usepackage{textcomp}%
\usepackage{manyfoot}%
\usepackage{booktabs}%
\usepackage{algorithm}%

\usepackage{algorithmicx}%
\usepackage{algpseudocode}%
\usepackage{listings}%
\usepackage{array}%
\usepackage{caption}
\usepackage{lscape}
\usepackage{tabularx}
\usepackage{adjustbox}%
\usepackage{subcaption} 
\usepackage{booktabs} 
\usepackage{multirow} 
\usepackage{hyperref} 
\newcolumntype{P}[1]{>{\centering\arraybackslash}p{#1}}

\begin{document}

\title[Artery-Vein Segmentation]{Artery-Vein Segmentation from
Fundus Images using Deep Learning}


\author[1]{\fnm{Sharan} \sur{SK}}\email{ced18i049@iiitdm.ac.in}
\equalcont{These authors contributed equally to this work.}

\author[2]{\fnm{Subin} \sur{Sahayam}}\email{ subinsahayamm@snuchennai.edu.in}
\equalcont{These authors contributed equally to this work.}

\author*[1]{\fnm{Umarani} \sur{Jayaraman}}\email{umarani@iiitdm.ac.in}

\author[3]{\fnm{Lakshmi Priya} \sur{A}}\email{priyaannamalai2002@gmail.com}

\affil*[1]{\orgdiv{Department of Computer Science and Engineering}, \orgname{Indian Institute of Information Technology Design and Manufacturing, Kancheepuram}, \orgaddress{ \city{Chennai}, \postcode{600127}, \country{India}}}

\affil[2]{\orgdiv{Department of Computer Science and Engineering}, \orgname{Shiv Nadar University}, \orgaddress{\city{Chennai}, \postcode{603110}, \country{India}}}

\affil[3]{\orgdiv{Department of Computer Science and Engineering}, \orgname{Amrita Vishwa Vidyapeetham}, \orgaddress{\city{Coimbatore}, \postcode{641112}, \country{India}}}


\abstract{Segmenting of clinically important retinal blood vessels into arteries and veins is a prerequisite for retinal vessel analysis. Such analysis can provide potential insights and bio-markers for identifying and diagnosing various retinal eye diseases. Alteration in the regularity and width of the retinal blood vessels can act as an indicator of the health of the vasculature system all over the body. It can help identify patients at high risk of developing vasculature diseases like stroke and myocardial infarction. Over the years, various Deep Learning architectures have been proposed to perform retinal vessel segmentation. Recently, attention mechanisms have been increasingly used in image segmentation tasks. The work proposes a new Deep Learning approach for artery-vein segmentation. The new approach is based on the Attention mechanism that is incorporated into the WNet Deep Learning model, and we call the model as Attention-WNet. The proposed approach has been tested on publicly available datasets such as HRF and DRIVE datasets. The proposed approach has outperformed other state-of-art models available in the literature.}




\maketitle

\section{Introduction}\label{sec1}

Eye diseases are prevalent and a growing cause for concern in the modern world. It is estimated that more than $2$ billion people worldwide are affected by visual impairments. More than half of the cases could have been prevented if these diseases were detected early (\cite{world2019world}). The majority of these diseases occur because of unhealthy lifestyles, and health-related problems such as smoking, drinking, ageing, and other systemic diseases. Most of the early signs are typically connected to defects in the shape and structure of the retinal blood vessels. 

Retinal vascular anomalies are crucial indicators for vasculature-related disorders such as hypertensive retinopathy, glaucoma,  myocardial infarction, and overall retinal vessel health (\cite{orlando2020refuge}). The overall number of people with eye illnesses and vision problems is expected to rise due to the combined effects of the increasing population and ageing.
Based on the survey results of \cite{vashist2022blindness}, more than one-fourth of persons aged $50$ years and above in India are visually impaired.

Fundus images are captured using a specialized fundus camera that gives features of a patient's retina. The image features include the retinal macula, optic disc, fovea, and blood vessels. These features make the fundus imaging method more suited for the non-invasive type of screening and are cost-effective. To track the development of specific eye diseases, optometrists, ophthalmologists, orthoptists, and other qualified medical experts utilize fundus photography for initial screening (\cite{ooi2021interactive}).  These images can be useful in scenarios such as emergencies where a patient has a diastolic pressure of $120$ mmHg or higher, patients with persistent headaches, patients with sudden visual loss, and even people under treatment for malaria (\cite{ooi2021interactive}).

\subsection{Need for Automatic Artery- Vein Segmentation}
Analyzing fundus images is an exhausting and tedious task even for trained experts as these images comprise complex components such as the structures of the retina and vascular system.  Retinal vascular segmentation and A/V categorization can help identify vessel health and predict early risk of diseases like stroke and heart attacks. However, because of the complexity of vessel structures, manually segmenting vessels is laborious and time-intensive, vulnerable to inter-rater variability, and has reduced efficiency (\cite{hu2021automatic}). Because of the limits of retinal image acquisition technologies, even specialists may find it difficult to identify such complex vessel structures in fundus images. Furthermore, the increasing volume of patient data adds to the challenge of clinical routines such as diagnosis, treatment, and monitoring (\cite{galdran2019uncertainty}). As a result, automated approaches for Artery-Vein segmentation from retinal images can be helpful in clinical settings.

Considering the degree of anatomical variation throughout the dataset and the possibility of inconsistent labelling among experts, automated categorization of retinal arteries and veins face computational difficulties. Most present approaches for measuring structural abnormalities suffer from arteriovenous confusion and blood vessel discontinuity (\cite{hu2022multi}). Identifying small arteries and veins can be especially challenging. Also, the class imbalance between the arteries and veins in the background region makes it difficult for representation-based learning models to generalize. Existing methods fail to distinguish blood vessels at bifurcations and crossover points as the 3D vision of the human eye is projected onto a 2D plane, which causes multiple different segments to emerge from the same vessel map. Moreover, vessels near crossover points are harder to distinguish as they look similar in appearance, shape, or contrast (\cite{zhao2019retinal}).

To address the problems explained so far, An automatic artery-vein classifier based on the Attention mechanism has been incorporated into W-Net (Encoder-Decoder) architecture \cite{xia2017wnet}. The model is capable of decreasing arterio-venous confusion by focusing on the foreground (minority) pixels that is arteries and veins rather than the background pixels.

\subsection{Gaps in Research}
Artery-Vein structures are complex and analyzing them requires the model to capture the morphological features and the topological connectivity of the vessel map. Graph-based methods capture the topological connectivity of the vessel but fail to provide accurate results because of unreliable or abrupt changes at junction points where two vessels meet. It makes it difficult to identify the presence of crossovers at these junctions and even a small error in labelling can propagate through the entire vessel structure (\cite{zhao2019retinal}).  Most of the Deep Learning models do not capture the topological connectivity and focus only on local features about a given point which fails at intersection points. Furthermore, existing methods fail to account for noisy label disturbance in limited datasets, and all of the methods mentioned above suffer from vessel identification discontinuities and arterio-venous confusion, particularly for smaller vessels and vessel intersections (\cite{hu2022multi}). Additionally, the majority of the public dataset has limited samples with ground truth (artery-vein blood vessels).
Also, the number of pixels for arteries and veins is imbalanced against the number of background pixels which reduces the performance of the trained models (\cite{li2020joint}).

The current state of retinal blood vessel analysis involves a variety of methods from semi-automatic techniques requiring manual input to advanced Machine Learning approaches using handcrafted features.  The current methodologies tend to fail for high inter-class similarity problems. Despite advancements, challenges persist such as the dependency on accurate segmentation maps, inconsistencies in image quality, and the inability of some models to handle complex vessel structures and noise effectively. \cite{estrada2014tree} Deep Learning models like U-Nets \cite{ronneberger2015unet} and attention mechanisms have been implemented to improve accuracy and efficiency, yet issues like over-fitting on small datasets and misclassification at vessel crossings remain significant hurdles. \cite{albadawi2018arterioles, hemelings2019artery}

\subsection{Contributions}
The important contribution of the work is highlighted here.
\begin{itemize}
\item A novel architectural approach using Attention-WNet for artery and vein segmentation in medical images has been proposed. The proposed approach achieved improved segmentation accuracy and robustness compared to conventional methods.

\item Unlike existing approaches that train on artery and vein segmentation together, the proposed strategy trains the network separately for artery and vein segmentation. It allows the model to learn distinct features and characteristics of both vessel types, leading to more accurate and meaningful segmentation results.

\item  The proposed approach integrates attention mechanisms, enabling the network to focus on relevant regions of interest while ignoring noise and irrelevant features. It leads to improved feature representation and helps the network ignore subtle differences between arteries and veins, contributing to higher segmentation accuracy. 

\item The proposed flow includes the use of focal loss to address class imbalance challenges commonly faced in artery-vein segmentation of fundus images. This approach helped to enhance segmentation performance by reducing overfitting of dominant vessel classes, thereby leading to more refined boundary delineations and improved overall accuracy.

\item

\end{itemize}

The rest of the paper is organized as follows. The related work is discussed in Section \ref{sec2}. The proposed approach is discussed in Section \ref{sec3}. Experimental results are presented in Section \ref{sec4}. The conclusion is given in Section \ref{sec5} and the references are in the last section.

\section{Related Work}\label{sec2}
Automatic evaluation of vascular abnormalities from fundus images depends on precise segmentation of retinal vessel pixels. When dealing with a large number of high-resolution sequences of images, manual segmentation becomes tedious. It consumes a large amount of time and is not feasible. As a result, various methods for automated artery and vein segmentation from fundus images have been introduced. The methods can be broadly classified into two categories i) traditional methods based on handcrafted features and ii) Deep Learning methods/models.

\subsection{Traditional Methods}
Traditional methods are semi-automatic methods that require a human expert to classify a branch as either a vein or an artery. \cite{martinez2002retinal} perform a semi-automatic retinal blood vessel analysis that is capable of measuring and quantifying the geometrical and topological properties. Early proposed methods are based on Machine Learning methods that incorporated handcrafted features to differentiate the artery and vein vessels from the background. Most of the traditional Machine Learning methods used color as a feature as it is a clear differentiating visual factor between the two. Due to the presence of excess oxygenated haemoglobin in the artery than in the veins. The latter looks darker in color compared to arteries, which led to the adoption of intensity-based criteria for vascular detection (\cite{yin2019artery}). Inconsistency in brightness and contrast often occurs in retinal images, both within and across images. These changes may have an impact on some approaches that rely on handcrafted features. Also, traditional methods often ignore the connectivity (cross-over pixels) of the vessels. Thus several graph-based methods have been introduced for solving artery/vein classification problems. 

Most of the Graph-based methods for artery vein classification require a separate vessel segmentation. It will be used for building the entire vascular network and to take advantage of the global information about the vessel structure. \cite{dashtbozorg2013automatic} proposed a method that splits the constructed graph into multiple sub-graphs that are later classified into artery/vein using an LDA classifier. \cite{estrada2014tree, estrada2015retinal} utilized a global likelihood model followed by a graph-theoretic method to capture the structural information of the blood vessel. \cite{rothaus2009separation} proposed an algorithm that does automatic graph separation. \cite{estrada2015retinal, de2017transduction} proposed a graph-theoretical approach to reconstruct the blood vessel network from topological information.

Since most of the methods are dependent on the vessel segmentation map, their accuracy is highly dependent on the correctness of the vessel map. Also, these methods employ features such as local growth, overlap, color, brightness, and hue to identify the artery-vein vessels present in the segmentation map. 
Some works in graph-based methods also use clustering techniques. \cite{de2017transduction} used a color-based clustering and blood vessel tracking strategy based on the minimal path approach. \cite{joshi2014automated, joshi2011automated} suggested a method that constructs subgraphs using Dijkstra’s shortest-path algorithm and labels each subgraph as either artery or vein using a fuzzy $K$-means clustering algorithm.

\subsection{Deep Learning Methods} 
With the recent development of Deep Learning methods in the field of medical imaging, many Deep Learning models have been proposed for solving artery vein classification problems. \cite{welikala2017automated} were the first to apply a Deep Learning model to recognize arteries and veins in fundus images without using traditional methods that involve handcrafted features. The proposed approach is based on Convolutional Neural Networks(CNNs) that have three convolutional and three fully connected layers. Since the model required centreline extraction and vessel segmentation, the model's output is based on the performance of vessel segmentation. Similar CNN-based models have been proposed by \cite{albadawi2018arterioles}, \cite{lepetit2018large}, and \cite{zhao2019retinal}. Girard's team trained a CNN for classifying pixels directly into an artery or veins.  Fraz's model is constructed based on the model architecture of SegNet \cite{badrinarayanan2017segnet}, which is an encoder-decoder-based fully connected CNN. Lepetit proposed a Large Receptive Field Fully CNN (LRFFCN). Lepetit's LRFFCN model can help in segmenting high-resolution images with reduced computational cost. Even in such cases, the proposed approach cannot learn the structural information of the vessel, which caused the model to perform subparly in the case of smaller vessels. \cite{wang2020multi} have proposed a multi-task Siamese network to learn more robust Deep Learning features about performing these two tasks together. 

State-of-the-art models for segmentation such as U-Net have also been proposed by \cite{hemelings2019artery} for solving A/V segmentation tasks combined with retinal vessel segmentation. But, even these models suffer from misclassification for smaller diameter A/V cross-over pixels. As the proposed approach is based on Encoder-Decoder Deep Learning architecture, it is useful in discussing UNet (\cite{ronneberger2015u}), WNet (\cite{galdran2020little}), and Attention UNet (\cite{oktay2018attention}) architectures.
 Hu et al. \cite{Hu2021} proposed Vessel-Constraint Network (VC-Net) addresses these issues with a vessel-constraint (VC) module and a multiscale feature (MSF) module. The VC module combines local and global vessel information, while the MSF module improves feature extraction across scales. Tested on various datasets, VC-Net demonstrated superior performance, achieving a balanced accuracy of 0.9554 and high F1 scores, proving its robustness and precision in A/V classification and vessel segmentation.

\subsection{UNet}
The U-Net architecture (\cite{ronneberger2015u}), introduced by Ronneberger, is a prominent convolutional neural network (CNN) designed for precise image segmentation, particularly in medical imaging. This "U"-shaped framework features a contracting path (encoder) with convolutional layers and max pooling for context capture, leading to a bottleneck. The expansive path (decoder) employs up-convolutions and concatenation to restore spatial resolution, aided by skip connections that preserve fine details. The final layer utilize $1 \times 1$ convolutions and activation functions to produce pixel-wise segmentation. U-Net excels in biomedical image analysis and can handle limited training data effectively. Its influence has extended to diverse segmentation tasks and advanced CNN architecture design.

\subsection{WNet} To enhance accuracy without compromising simplicity, a direct alteration is employed to the U-Net architecture, which is called WNet.
WNet (\cite{galdran2020little}) Architecture is a combination of two UNets combined sequentially, which is defined by the equation: 
\begin{equation}
        \theta(x) = \phi^2(x,\phi^1(x))
\end{equation}
where, $x$ denotes the input image, $\phi^1(x)$ 
generates the first prediction of artery-vein segmentation that can then be used by $\phi^2(x)$ as a sort of attention mechanism to focus more on interesting areas of the retinal image
and $\theta(x)$ denotes the final output from the WNet architecture.
In general, WNet consists of two $\phi_{3,8}$ UNets. Hence, the number of parameters involved in a WNet is roughly more than $68000$ weight parameters, which is still lesser than other complex Deep Learning models for artery-vein segmentation. 
MSGANet-RAV, introduced by A Z M Ehtesham Chowdhury et al. \cite{Chowdhury2022} utilizes a U-shaped encoder-decoder network. This design extracts multiscale features and incorporates self-attention modules, enhancing the automation of complex retinal vessel structure labelling, a task traditionally prone to complications in manual methods.

\subsection{Attention UNet }
The Attention UNet (\cite{oktay2018attention}) is an extension of the popular U-Net architecture, which is a fully convolutional neural network for biomedical image segmentation, that incorporates attention mechanisms to selectively focus on informative regions of the feature maps. The Attention-UNet extends the UNet architecture by adding an attention mechanism that selectively focuses on the most informative regions of the feature maps. The attention mechanism is applied to the skip connections between the encoder and decoder networks, allowing the network to learn where to focus its attention for better segmentation results. The attention mechanism is achieved through a mechanism called the "Squeeze-and-Excitation" (SE) block, which learns to scale the feature maps based on their importance.

The Attention U-Net architecture starts with an input image and processes it through an encoder path that consists of convolutional blocks with ReLU activations and max-pooling operations. This sequence of operations progressively reduces the spatial dimensions of the feature maps, capturing detailed contextual information at multiple levels. Attention gates are incorporated into the skip connections between the corresponding layers of the encoder and decoder paths. These gates use the feature maps from the encoder and decoder to create attention-modulated maps that highlight the most relevant regions. The decoder path then reconstructs the segmentation map by up-sampling the feature maps and concatenating them with the attention-modulated encoder maps. This concatenation ensures that spatial details are preserved and integrated with the higher-level features. The combined feature maps undergo further convolutional processing to refine the segmentation. By using Attention Gates in the skip connections, the model selectively focuses on important features, improving the accuracy of the final high-resolution segmentation map, which accurately highlights regions of interest in the input image.

\subsection{Attention WNet}
The Attention W-Net model \cite{mitta2020upgraded} incorporates attention gates into the skip connections of a W-Net (dual U-Net) structure, enabling it to focus more effectively on relevant regions and reduce segmentation noise. To optimize unsupervised 3D segmentation, the model uses a soft N-Cut loss and Structural Similarity Index (SSIM) loss, with Conditional Random Fields (CRFs) added in post-processing to refine the segmentation.
\section{Proposed Model}\label{sec3}
\begin{figure}[htb]
\centering
  \includegraphics[width=1\textwidth]{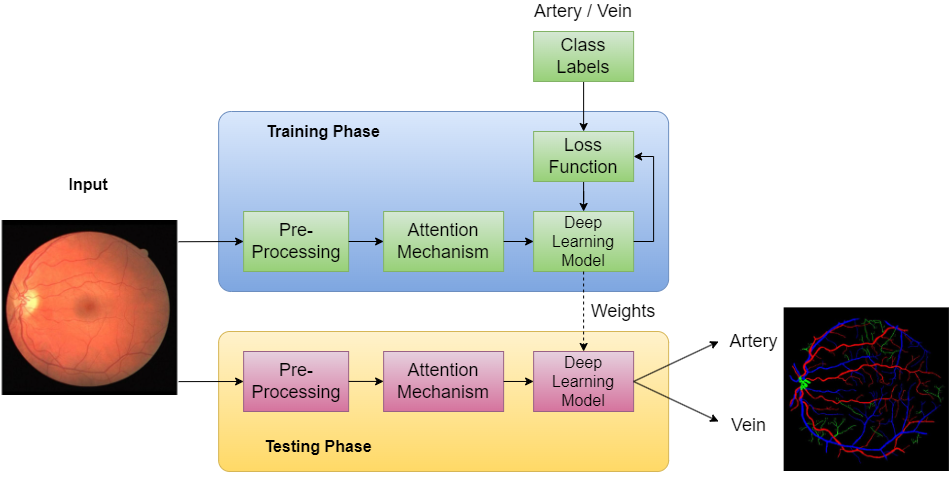}
  \caption{The flow Diagram of the Proposed Model}
  \label{fig: Block}
\end{figure}

The block diagram of the proposed model is shown in Fig. \ref{fig: Block}. The proposed model pipeline includes several important components such as input and pre-processing, attention mechanism, attention-WNet Deep Learning model, and the loss function model for segmentation. Each of these components is explained below.

\subsection{Model Pipeline}
The utilized model sometimes suffers from low-range inter-class variability between arteries and vessels which can be visualized for an HRF sample in Figure \ref{fig:Image_Histogram}. It might be because of the lower contrast between the artery, vein, and background (\cite{soares2006retinal}) or the model is biased towards the presence of more artery pixels as compared to veins (\cite{staal2004automated}). To overcome these difficulties, a new approach has been proposed.

In contrast to existing approaches that train on artery and vein segmentation together, this work proposes training the network separately for artery segmentation and separately for vein segmentation. This enables the model to capture the unique features and characteristics of the artery and the vein separately, leading to more accurate and meaningful segmentation results. To better differentiate between arteries and veins, the method utilizes two separate Attention-WNet models, each dedicated to one vessel type. The problem becomes a less complex binary segmentation task, thereby reducing the complexity of the model.

\subsection{Input and Pre-Processing}
\begin{figure}[htb]
\centering
  \includegraphics[width=1\textwidth]{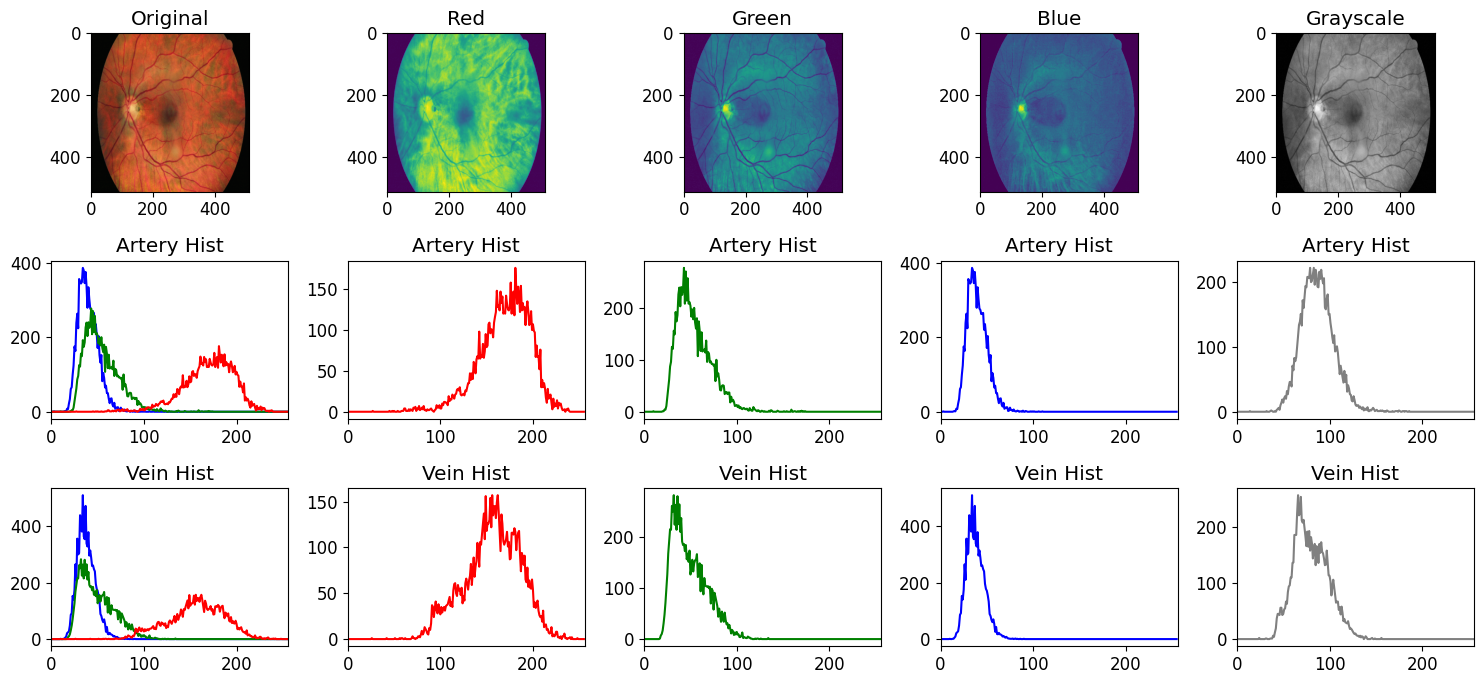}
  \caption{Channel-wise histogram of artery and veins for original, red, green, blue, and grayscale fundus images of a single patient.  }
  \label{fig:Image_Histogram}
\end{figure}

The proposed flow is trained on two datasets, namely DRIVE and HRF. Fig. \ref{fig:Image_Histogram} shows a sample from DRIVE and HRF images and the variability of pixel values across different channels. To improve the cross-dataset performance, z-score normalization is performed to standardize the pixel intensity values across both datasets. Due to the large size of HRF, training the model on a T4 or P100 will be difficult, hence the dataset is resized to $1024 \times 1024$ for HRF and $512 \times 512$ for DRIVE.

\begin{figure}[htb]
   \includegraphics[width=1\textwidth]{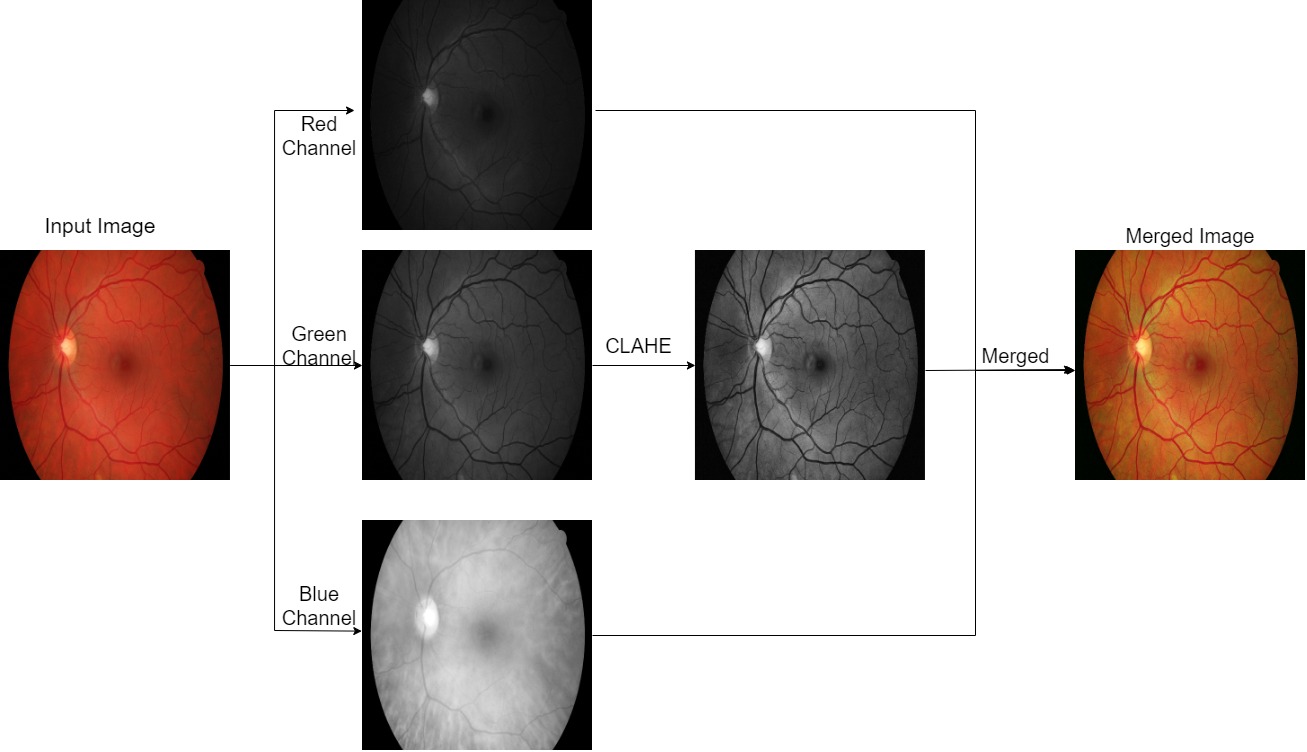}
  \caption{Image enhancement through CLAHE on the green channel from the input fundus image(\cite{ronneberger2015u})}
  \label{fig:Image_Clage}
\end{figure}

It is observed from Fig. \ref{fig:Image_Histogram} that among all channels, green channels show better contrast between arteries and veins and it can be improved by applying Contrast-Limited Adaptive Histogram Equalization (CLAHE) operation (\cite{saleh2011automated}).  The images are enhanced by applying CLAHE only to the green channel of the image as shown in figure \ref{fig:Image_Clage}.

\subsection {Attention Mechanism }
The attention mechanism is incorporated into standard WNet architecture to highlight salient features that are passed through the skip connections. The attention mechanism distinguishes the irrelevant and noisy response at skip connection. The mentioned step is performed before the concatenation operation to combine only relevant activation. In addition, the attention mechanism filters the activation during forward and backward passes. It allows only relevant layers to be updated that are useful for the segmentation task. The update rule for convolution parameters in layer $l-1$ can be formulated as follows:

\begin{equation}
\frac{\partial (\hat{x_i}^l)}{\partial (\phi ^{l-1})} = \frac{\partial (\alpha_i^l f (x_i^{l-1};\phi ^{l-1})}{\partial (\phi ^{l-1})} = \frac{\alpha_i^l \partial ( f (x_i^{l-1};\phi ^{l-1}))}{\partial (\phi ^{l-1})}+
\frac{\partial (\alpha_i^l )}{\partial (\phi ^{l-1})} x_i^l
\end{equation}

The first gradient term on the right-hand side is scaled with  $\alpha_i^l$,  In the case of multi-dimensional attention mechanism, $\alpha_i^l$ corresponds to a vector at each grid scale.
 Here, deep supervision is used to have discriminative feature maps at each image scale. This ensures the attention unit to get responses from foreground content rather than background content
 Therefore, it prevents dense predictions from being reconstructed
from small subsets of skip connections.

\subsection{Attention-WNet Deep Learning Model}
Although WNet is an effective model for segmentation, the model still suffers from its inability to generalize with new datasets and capture the long-term dependencies of artery vein structure. To resolve these issues, a model has been utilized, that combines the Attention mechanism  \cite{oktay2018attention}) with WNet architecture called the Attention-WNet model \cite{mitta2020upgraded}. It is used to selectively focus key characteristics in the input image while ignoring irrelevant features. Attention mechanisms are useful in the context of artery-vein segmentation from fundus images to help recognize the essential characteristics that distinguish arteries from veins while ignoring the features that are common in both.

\begin{figure}[H]
\rotatebox[origin=c]{270}{\includegraphics[width=1.5\textwidth]{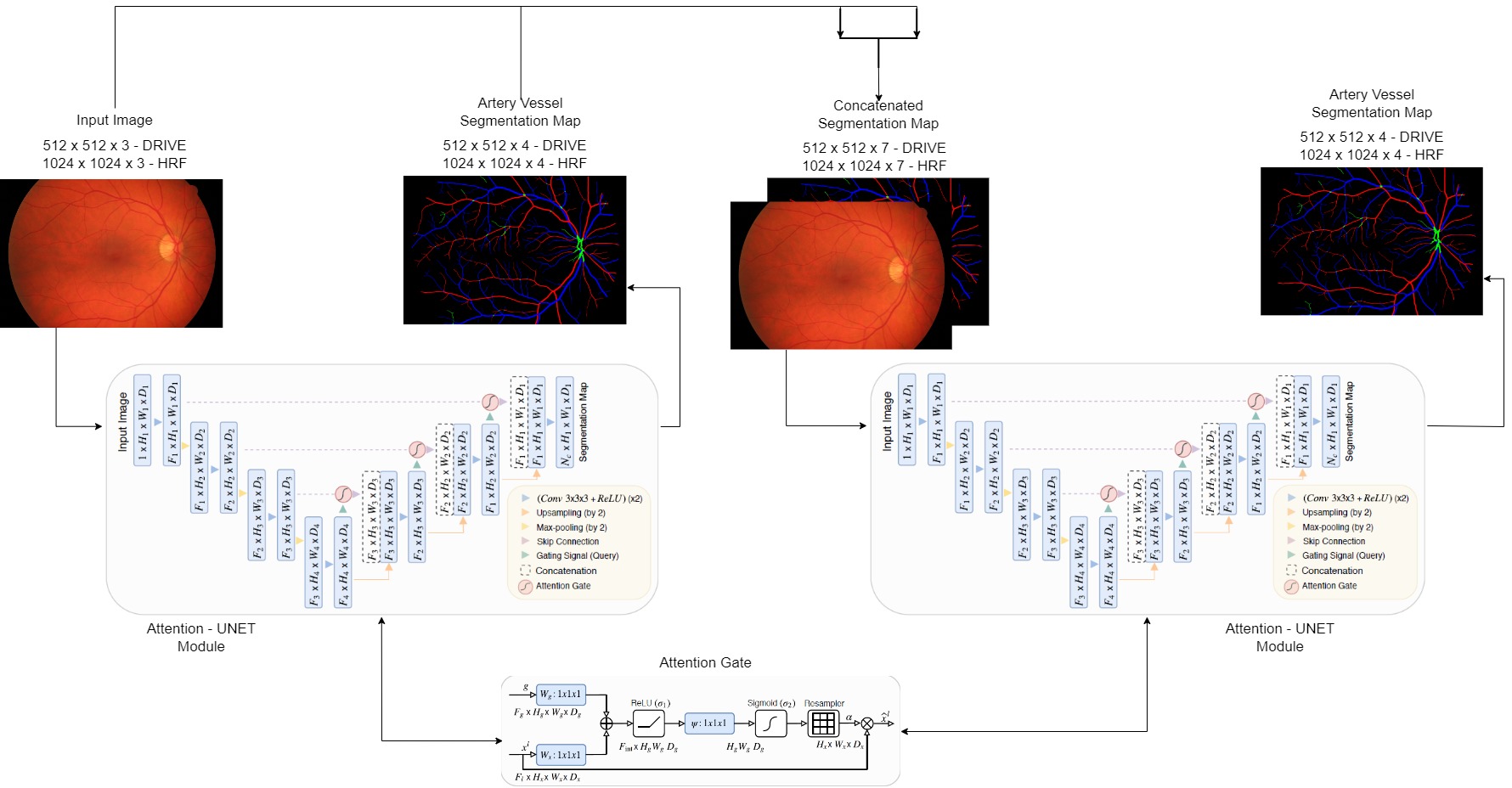}}
  \caption{Proposed approach using Attention-WNET Deep Learning Model for AV segmentation }
  \label{fig: AttentionWNet}
\end{figure}

In the proposed flow, attention mechanisms are added to the skip connections of the first Attention-UNet block by passing the input image, $x$. It helps to improve the features that are essential for segmentation while ignoring the irrelevant features. The attention mechanisms can be trained to learn the weights that assign value to the various features in the input image. The output obtained from the first Attention-UNet block is then passed to the second Attention-UNet block along with the input image, $x$.
It improves the features of the input image and extracts more complex features. The Attention-UNet blocks are made up of a sequence of convolutional layers and an attention mechanism that is capable of capturing the local and global dependencies between image features. The Attention-UNet block's predicted output is then passed into the WNet decoder, where the segmentation result is obtained. Each of the Attention-UNet blocks predicts an artery vein segmented image, as explained in Section 4.3. The architectural diagram of the Attention-WNet model for artery vein segmentation is shown in Fig. \ref{fig: AttentionWNet}.

\subsection{Loss Function}
Since the majority of the pixels are background, the arteries and veins belong to minority classes, the utilized model has a high chance of suffering from class imbalance problems. Focal loss (\cite{lin2017focal}) is used to improve the performance of a model concerning minor classes by reducing the relative loss for correctly classified pixels and focusing more on misclassified pixels. The main advantage of using focal loss is that it does not affect the performance of the background class while improving the artery and vein performance. The formula for calculating focal is given in equation \ref{eq:focal_loss}, where $i$ corresponds to ground truth class label for the $i$th class, $p_i$ is the predicted probability for the $i$th class and $p_t$ is the predicted probability for the true class label; $\alpha_t$ is a weighting factor that is applied to the loss for the true class and is typically set to $(1 - \beta_t)$, where $\beta_t$ is the true class frequency; $\gamma$ is the focusing parameter that controls the rate at which the loss of correctly-classified pixels is down-weighted relative to misclassified pixels. Focal loss is used here with a $\gamma$, gamma value of $2$, an $\alpha$, alpha value of $0.8$ for artery, and vein, and $0.9$ for learning uncertain pixels.

\begin{equation}
\label{eq:focal_loss}
FL(p_t) = -\alpha_t (1 - p_t)^\gamma \sum_{i=1}^{C}y_i \log(p_i),
\end{equation}

\section{Experimental Results}\label{sec4}
The section details the various datasets used in evaluating the study. It is followed by the performance metrics used to measure the performance of the proposed approach. The later subsections discuss the results obtained and the comparison between the proposed methods and the other methods in the literature.

\subsection{Datasets}
The datasets used in the training are Drive-AV and HRF datasets. Both DRIVE and HRF Datasets contain a small number of fundus images for training a U-Net-based model (\cite{shen2018retinal}). DRIVE contains $40$ images of size $565 \times 584$ pixels, whereas HRF contains $45$ images of size $3504 \times 2336$ pixels. The details of the dataset that are provided in the official source are shown in table \ref{tab: Dataset_Description}. For our experimentation, we have considered splitting the dataset into train and test sets, where, each dataset contains approx of $30$ images for training based on an $80:20$ split.

\begin{table}[tbh]
\caption{Dataset Description}
\label{tab: Dataset_Description}
\begin{tabular}{llll}
\hline
\begin{tabular}[c]{@{}c@{}}\textbf{Dataset}\\ \textbf{Name}\end{tabular} & \begin{tabular}[c]{@{}c@{}}\textbf{Image}\\ \textbf{Resolution}\\ \textbf{(Width x Height)}\end{tabular} & \begin{tabular}[c]{@{}c@{}}\textbf{No of Images}\\ \textbf{in the Dataset}\end{tabular}                                                                                                                                                                                                                                                              & \begin{tabular}[c]{@{}c@{}}\textbf{Dataset}\\ \textbf{Description}\end{tabular}                                                                                                                                                                                                                                                                                                                                                                                                  \\ \hline
\begin{tabular}[c]{@{}l@{}}DRIVE -\\ AV\end{tabular}   & 565 x 584                                                                     & \begin{tabular}[c]{@{}l@{}}40 Images:\\ Training - 20 images\\ Test - 20 images\end{tabular}                                                                                                                                                       & \begin{tabular}[c]{@{}l@{}}Each image consists of :\\ - Binary Mask for FOV Area.\\ - Pixel Wise Labeling for:\\      -Vessel Segmentation\\      - A/V Classification\\ \\ A/V Classification has four\\ types of labels:\\ - Arteries are labeled in red.\\ - Veins are labeled in blue.\\ - Cross-over points (overlapping\\ of arteries and veins and uncertain\\ pixels) are labeled in green.\\ - Background pixels are labeled\\ in black.\end{tabular} \\ \hline
HRF                                                    & 3504 x 2336                                                                   & \begin{tabular}[c]{@{}l@{}}45 Images:\\ - The dataset is split into\\ three categories:\\      - Healthy\\      - Diabetic Retinopathy\\      - Glaucoma\\ - Training: 30 images - \\           15 images from each\\            category\\ - Testing: 15 images -\\          5 images from each\\           category\end{tabular} & Same as DRIVE - AV                                                                                                                                                                                               \\ \hline
\end{tabular}
\end{table}

\subsection{Performance Metrics}
Accuracy and Dice Coefficient (Dice / F1-Score) are used to measure the performance of the architectures in artery-vein classification.
\begin{equation}
        F1-Score =  \frac{2*TP}{2*TP + FP + FN}
\end{equation}
\begin{equation}
         Accuracy =  \frac{TP + TN}{TP + FP + FN + TN} 
\end{equation}
where the definition of TP, TN, FP, and FN for the artery class is defined below and this definition is extended for vein, background, and uncertain pixel classes.

\begin{itemize} \item True Positive (TP): The number of pixels correctly classified as arteries. \item True Negative (TN): The number of pixels correctly classified as background, vein, or uncertainty. \item False Positive (FP): The number of pixels incorrectly classified as arteries when they actually belong to the background, vein, or uncertainty. \item False Negative (FN): The number of pixels incorrectly classified as background, vein, or uncertainty when they actually belong to arteries. \end{itemize}

\subsection{Experimental Setup}

In this paper, UNet architecture is characterized by $\phi_{k,f_0}$, where, $k$ denotes the number of times the image has been upscaled or downscaled (depth levels) and $f_0$ denotes the number of filters used in each of the depth levels.
To simplify the  analysis, filters
of size, $3 \times 3$ are considered, and the number of filters is doubled each time as $k$ increases. The base UNet architecture that will be used in the paper is the $\phi_{3,8}$ version. The model has extra skip connections within each block and Batch-Normalization is applied after each convolution operation. The model contains around $34000$ weight parameters(approx) and is still lesser than the parameters employed in other CNN models that are used in retinal vessel segmentation.

The model has been trained using an A100 GPU with 40 GB of VRAM and 64 GB of memory. Since the proposed approach is a computationally intensive model for an input image resolution of 1024x1024 and considering the available resources, the batch size was limited to 4, 6, and 8, and the epochs were increased to 200 to compensate for the smaller batch size, followed by early stopping with a patience value of 20 to find the optimal results. Finally, the models have been trained against DRIVE and HRF datasets for 200 epochs, with a batch size of 6, and using Adam Optimizer with a learning rate of 0.01. HRF dataset images are of 3504 × 2336 pixels and DRIVE is of 564x585 pixels. It is converted to an image size of 1024x1024 and 512x512, respectively. In the case of cross-dataset verification, the image resolution has been upscaled or downscaled to the resolution of the dataset that the model was originally trained on. 

Some of the pixels might not have a clear classification as either an artery or a vein. Such uncertainty may arise from diverse factors, such as minimal contrast between arteries and veins, image noise, indistinct features, or pathological conditions altering typical vascular structures. Owing to the presence of these uncertain pixels,  the output of the model will have four classes, namely, arteries, veins, uncertain regions, and other background pixels. A pixel will be marked based on the output probability for each pixel. In case, model 1 predicts a pixel as an artery and model 2 predicts the same pixel as a vein, then that corresponding pixel will be marked as an artery or vein based on whichever model has a higher activation value for that pixel.

In case, the activation values of the output are similar and within a difference of 20\%, then it will be marked as an uncertain pixel, suggesting that the pixel shares similar characteristics of both traits.

\subsection{Evaluation Procedure}
\begin{figure}[H]
\hspace*{-0.8cm}
\includegraphics[width=1.1\textwidth, trim={0 0 0 3cm}, clip]{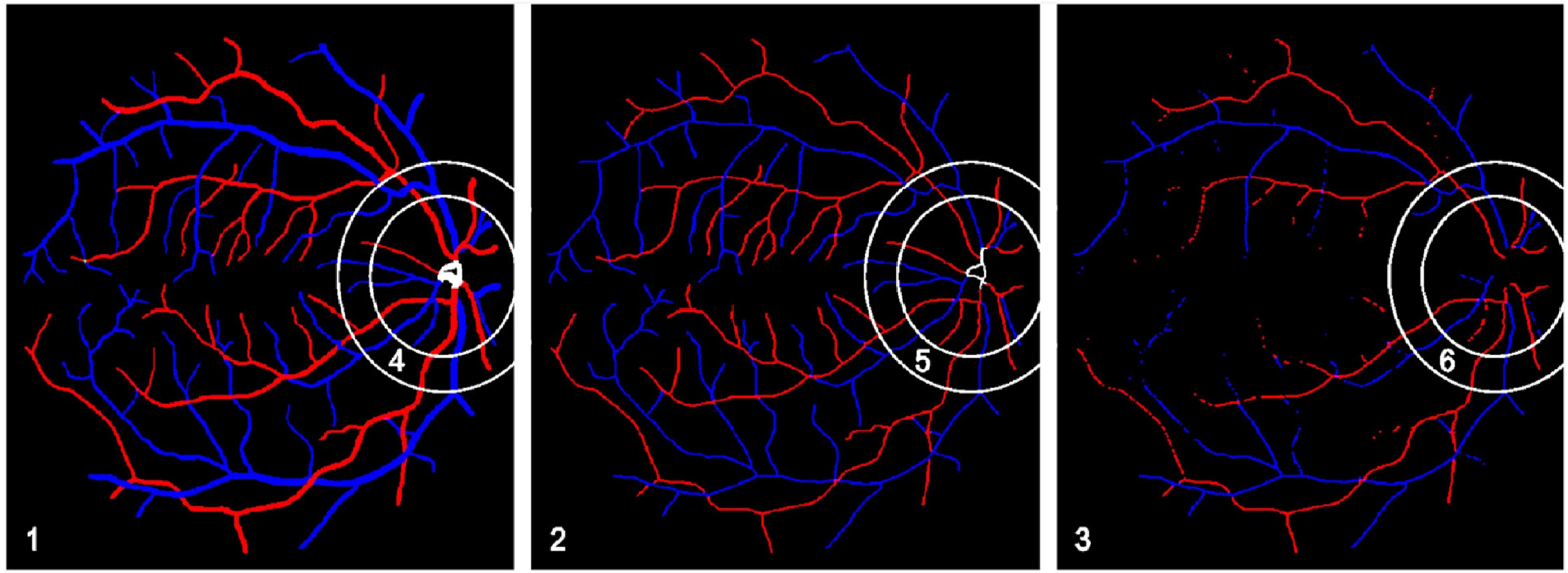}
  \caption{Evaluation Procedure: Illustration of the different metrics used in the evaluation of A/V segmentation, from left to right (1) All vessel pixels (2) Vessel centerline pixels (3)) Centerline pixels  of vessels wider than two pixels (\cite{hemelings2019artery}) }
  \label{fig: Evaluation_flow}
\end{figure}

Evaluation of the proposed approach is carried out on test sets of DRIVE and HRF datasets. From many of the existing works on artery-vein segmentation, it can be observed that the accuracy and F1-Score are often 
high when compared pixel-to-pixel for the whole predicted output, due to the high number of background pixels. The proposed method follows the evaluation metrics of (\cite{hemelings2019artery}). The evaluation involves calculating accuracy and F1-score across different regions of the predicted output. First, the metrics are calculated for the entire output, Second for the centerline vessel pixels of width equal to $1$ pixel, and Third for the centerline vessel pixels of width greater than $1$ pixel. It is helpful to conduct a comprehensive study of how the model performs over varied thicknesses. For example, results might be high when we compared the whole image, but when the results are compared for vessels with thickness greater than or equal to 1, the model's weakness in handling varied pixel widths can be identified. A sample of the evaluation method is shown in Fig. \ref{fig: Evaluation_flow}.

\subsection{Results and Analysis}

Segmenting arteries and veins is a tedious task. From the previous sections, it is well known that the inter-class variability (\cite{kirbas2004review}) is very low and that often model classifies a pixel as an artery instead of a vein. Observing the individual image results, most predicted vessels are of artery class. It is possible due to the less prominent nature of veins as compared to arteries (\cite{li2019automatic}). Veins are generally larger but are thinner vessels than arteries. It makes them less visible as they can be compressed. Additionally, due to the lower pressure present in the vein as compared to arteries, the contrast between the background pixels and veins is bound to decrease (\cite{soares2006retinal}).

\begin{table}[htb]
\centering
\caption{Results of Artery/Vein Classification on HRF Dataset}
\label{tab:Binary_HRF_Dataset}
\begin{tabular}{llllllll}
\hline
Model                                                                          &  & \multicolumn{2}{l}{\begin{tabular}[c]{@{}l@{}}All vessel\\ pixels\end{tabular}}                                & \multicolumn{2}{l}{\begin{tabular}[c]{@{}l@{}}Discovered vessel\\ centerline pixels\end{tabular}}             & \multicolumn{2}{l}{\begin{tabular}[c]{@{}l@{}}Discovered centerline\\ pixels of vessels wider\\ than two pixels\end{tabular}} \\ \hline
                                                                               &  & Accuracy                                               & F1                                                    & Accuracy                                              & F1                                                    & Accuracy                                                      & F1                                                            \\ \hline
\begin{tabular}[c]{@{}l@{}}Swin-WNet\\ (Seperate A-V)\end{tabular}             &  & \begin{tabular}[c]{@{}l@{}}0.916 ${\pm}$\\ 0.023\end{tabular} & \begin{tabular}[c]{@{}l@{}}0.884 ${\pm}$\\ 0.019\end{tabular} & \begin{tabular}[c]{@{}l@{}}0.730 ${\pm}$\\ 0.015\end{tabular} & \begin{tabular}[c]{@{}l@{}}0.721 ${\pm}$\\ 0.021\end{tabular} & \begin{tabular}[c]{@{}l@{}}0.743 ${\pm}$\\ 0.018\end{tabular}         & \begin{tabular}[c]{@{}l@{}}0.722 ${\pm}$\\ 0.016\end{tabular}         \\ \hline
Swin-WNet                                                                      &  & \begin{tabular}[c]{@{}l@{}}0.948 ${\pm}$\\ 0.013\end{tabular}  & \begin{tabular}[c]{@{}l@{}}0.937 ${\pm}$\\ 0.016\end{tabular} & \begin{tabular}[c]{@{}l@{}}0.712 ${\pm}$\\ 0.098\end{tabular} & \begin{tabular}[c]{@{}l@{}}0.711 ${\pm}$\\ 0.104\end{tabular} & \begin{tabular}[c]{@{}l@{}}0.718 ${\pm}$\\ 0.100\end{tabular}         & \begin{tabular}[c]{@{}l@{}}0.718 ${\pm}$\\ 0.106\end{tabular}         \\ \hline
Attention-UNet                                                                 &  & \begin{tabular}[c]{@{}l@{}}0.933 ${\pm}$\\ 0.009\end{tabular}  & \begin{tabular}[c]{@{}l@{}}0.930 ${\pm}$\\ 0.041\end{tabular} & \begin{tabular}[c]{@{}l@{}}0.708 ${\pm}$\\ 0.039\end{tabular} & \begin{tabular}[c]{@{}l@{}}0.691 ${\pm}$\\ 0.035\end{tabular} & \begin{tabular}[c]{@{}l@{}}0.726 ${\pm}$\\ 0.030\end{tabular}         & \begin{tabular}[c]{@{}l@{}}0.713 ${\pm}$\\ 0.028\end{tabular}         \\ \hline
\textbf{Attention - WNet}                                                               &  & \begin{tabular}[c]{@{}l@{}}\textbf{0.960} ${\pm}$\\ \textbf{0.005}\end{tabular}  & \begin{tabular}[c]{@{}l@{}}\textbf{0.958} ${\pm}$\\ \textbf{0.007}\end{tabular} & \begin{tabular}[c]{@{}l@{}}\textbf{0.810} ${\pm}$\\ \textbf{0.036}\end{tabular} & \begin{tabular}[c]{@{}l@{}}\textbf{0.810} ${\pm}$\\ \textbf{0.037}\end{tabular} & \begin{tabular}[c]{@{}l@{}}\textbf{0.845} ${\pm}$\\ \textbf{0.037}\end{tabular}         & \begin{tabular}[c]{@{}l@{}}\textbf{0.845} ${\pm}$\\ \textbf{0.049}\end{tabular}         \\ \hline

\end{tabular}
\end{table}

\begin{table}[htb]
\centering
\caption{Results of Artery/Vein Classification on DRIVE Dataset}
\label{tab:Binary_DRIVE_Dataset}
\begin{tabular}{llllllll}
\hline
Model                                                                        &  & \multicolumn{2}{l}{\begin{tabular}[c]{@{}l@{}}All vessel\\ pixels\end{tabular}}                                & \multicolumn{2}{l}{\begin{tabular}[c]{@{}l@{}}Discovered vessel\\ centerline pixels\end{tabular}}             & \multicolumn{2}{l}{\begin{tabular}[c]{@{}l@{}}Discovered centerline\\ pixels of vessels wider\\ than two pixels\end{tabular}} \\ \hline
                                                                             &  & Accuracy                                               & F1                                                    & Accuracy                                              & F1                                                    & Accuracy                                                      & F1                                                            \\ \hline
\begin{tabular}[c]{@{}l@{}}Swin-WNet\\ (Seperate A-V)\end{tabular}           &  & \begin{tabular}[c]{@{}l@{}}0.943 ${\pm}$\\ 0.007\end{tabular} & \begin{tabular}[c]{@{}l@{}}0.932 ${\pm}$\\ 0.012\end{tabular} & \begin{tabular}[c]{@{}l@{}} \textbf{0.778} ${\pm}$\\ \textbf{0.008}\end{tabular} & \begin{tabular}[c]{@{}l@{}}0.694 ${\pm}$\\ 0.006\end{tabular} & \begin{tabular}[c]{@{}l@{}}\textbf{0.791} ${\pm}$\\ \textbf{0.013}\end{tabular}         & \begin{tabular}[c]{@{}l@{}}0.757 ${\pm}$\\ 0.017\end{tabular}         \\ \hline
Swin-WNet                                                                    &  & \begin{tabular}[c]{@{}l@{}}0.903 ${\pm}$\\ 0.019\end{tabular}  & \begin{tabular}[c]{@{}l@{}}0.919 ${\pm}$\\ 0.010\end{tabular} & \begin{tabular}[c]{@{}l@{}}0.582 ${\pm}$\\ 0.011\end{tabular} & \begin{tabular}[c]{@{}l@{}}0.580 ${\pm}$\\ 0.014\end{tabular} & \begin{tabular}[c]{@{}l@{}}0.649 ${\pm}$\\ 0.015\end{tabular}         & \begin{tabular}[c]{@{}l@{}}0.647 ${\pm}$\\ 0.016\end{tabular}         \\ \hline
Attention-UNet                                                               &  & \begin{tabular}[c]{@{}l@{}}0.948 ${\pm}$\\ 0.005\end{tabular}  & \begin{tabular}[c]{@{}l@{}}0.941 ${\pm}$\\ 0.008\end{tabular} & \begin{tabular}[c]{@{}l@{}}0.766 ${\pm}$\\ 0.027\end{tabular} & \begin{tabular}[c]{@{}l@{}}\textbf{0.765} ${\pm}$\\ \textbf{0.027}\end{tabular} & \begin{tabular}[c]{@{}l@{}}0.774 ${\pm}$\\ 0.027\end{tabular}         & \begin{tabular}[c]{@{}l@{}}0.730 ${\pm}$\\ 0.027\end{tabular}         \\ \hline
\textbf{Attention - WNet}                                                             &  & \begin{tabular}[c]{@{}l@{}}\textbf{0.950} ${\pm}$\\ \textbf{0.004}\end{tabular}  & \begin{tabular}[c]{@{}l@{}}\textbf{0.946} ${\pm}$\\ \textbf{0.005}\end{tabular} & \begin{tabular}[c]{@{}l@{}}0.764 ${\pm}$\\ 0.041\end{tabular} & \begin{tabular}[c]{@{}l@{}}0.762 ${\pm}$\\ 0.042\end{tabular} & \begin{tabular}[c]{@{}l@{}}0.780 ${\pm}$\\ 0.042\end{tabular}         & \begin{tabular}[c]{@{}l@{}}\textbf{0.778} ${\pm}$\\ \textbf{0.042}\end{tabular}         \\ \hline

\end{tabular}
\end{table}


Tables \ref{tab:Binary_HRF_Dataset} and \ref{tab:Binary_DRIVE_Dataset} show the results obtained from training and validation within the HRF and DRIVE datasets, respectively. The experimentation results show that Attention-WNet performs on par with state-of-the-art models for artery-vein segmentation tasks. The consistently improved performance across all metrics shows the Attention WNet learns better and more accurate features compared to the other architectures in the literature.

\begin{figure}[H]
    \centering
    \subfloat[Fundus Image from DRIVE Dataset\label{gco_a}]
    {\includegraphics[width=0.4\linewidth]{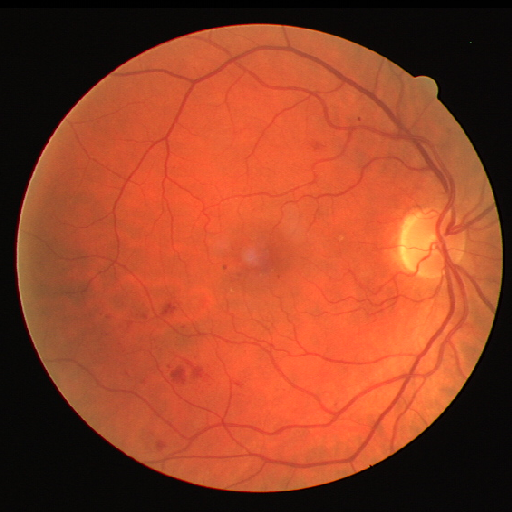}}
    \quad
    \subfloat[Ground Truth for the Fundus Image\label{gco_b}]
    {\includegraphics[width=0.4\linewidth]{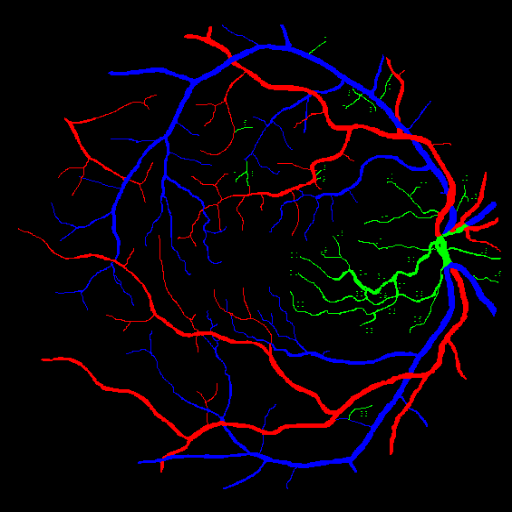}}
 
    \subfloat[Attention UNet's Predicted output \label{gco_c}]
    {\includegraphics[width=0.4\linewidth]{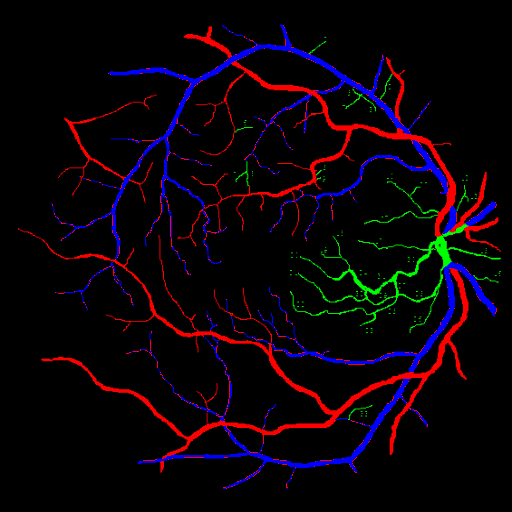}}
    \quad
    \subfloat[Attention WNet's Predicted output \label{gco_d}]
    {\includegraphics[width=0.4\linewidth]{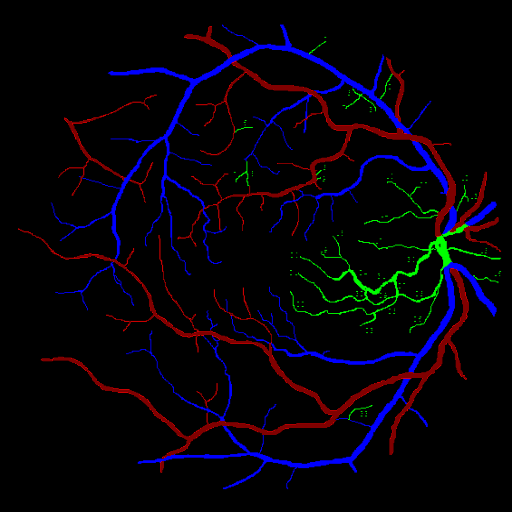}}
    \caption{Comparing Attention UNet and Attention WNet output for a given fundus image}
    \label{fig:gt2}    

\end{figure}

\begin{figure}[H]
\centering
\includegraphics[width=1\textwidth]{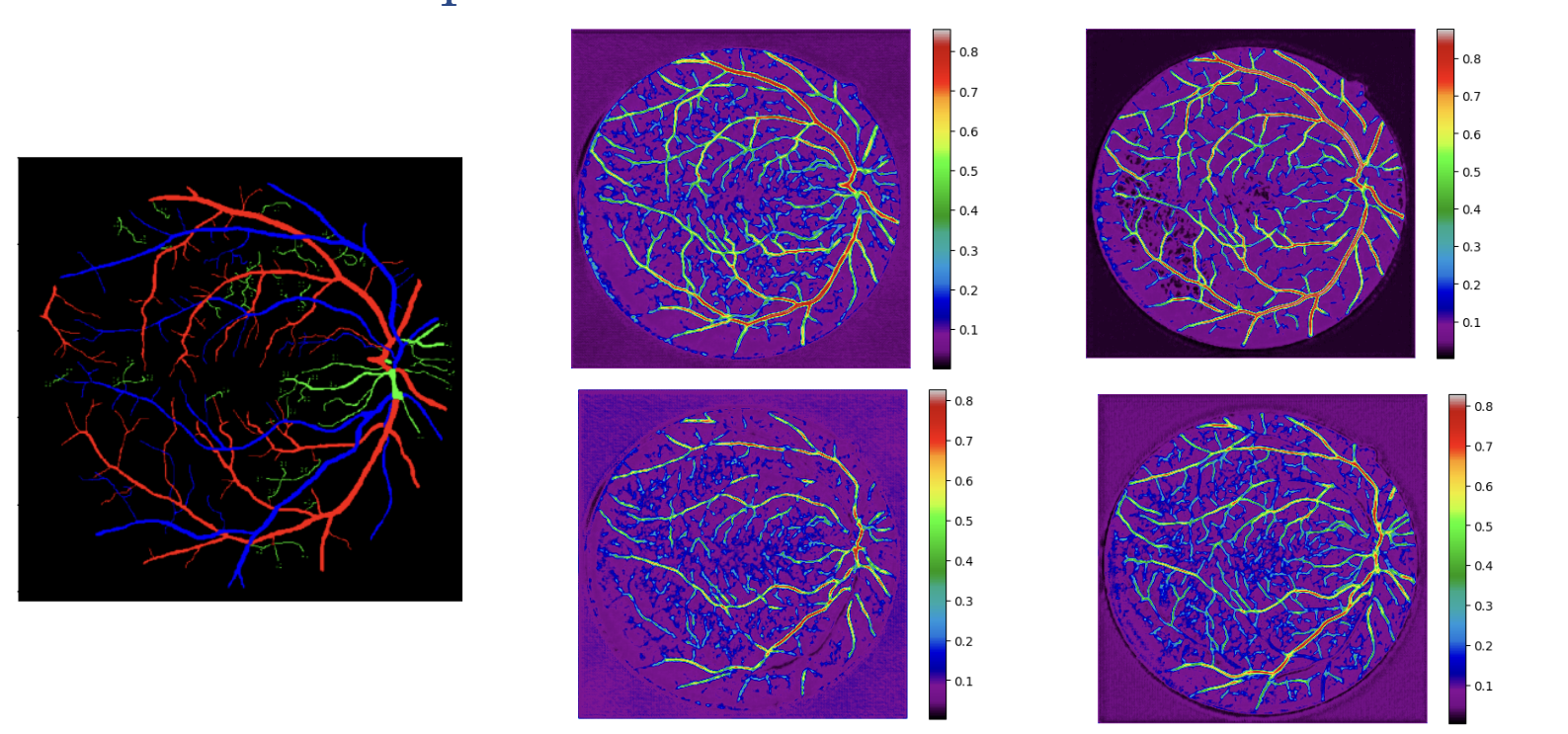}
  \caption{Activation Map for Model trained using Separate Artery-Vein segmentation and Binary Segmentation }
\label{fig: Activation_Map}
\end{figure}

For better explainability of the result, Tables \ref{tab:Binary_HRF_Dataset_new} and \ref{tab:Binary_DRIVE_Dataset_new} and the activation map of the final layer can be observed. In Fig. \ref{fig: Activation_Map}, the first image corresponds to the ground image, the first column corresponds to the activation map of the artery and vein channel of the Attention-UNet Model trained for Separate Artery-Vein segmentation, and the second column corresponds to the activation map of artery and vein channel of the Attention-WNet Model trained for binary segmentation model. It can be observed that the latter model is more confident about the vessel's nature than the former.
The fundus image, ground truth, and the corresponding sample outputs provided by Attention UNet and Attention-WNET are shown in Figure \ref{fig:gt2}. It can be observed that some of the vein pixels have been marked as artery pixels by Attention UNet. In this case, Attention WNet predicts better than Attention UNet.
It shows the proposed approach's capability to distinguish between arteries and veins. It is also observed that the model is comparatively more confident about artery pixels than veins, further confirming the complexity involved in identifying veins even though the latter performs comparatively better. 

\begin{table}[htb]
\caption{Results of Artery/ Vein Classification on HRF Dataset with Separate Metrics}
\label{tab:Binary_HRF_Dataset_new}
\begin{tabular}{llllll}
\hline
\textbf{Model}       &       & \multicolumn{2}{l}{\textbf{Artery Accuracy}} & \multicolumn{2}{l}{\textbf{Vein Accuracy}} \\ \hline
Attention-UNET       &       & \multicolumn{2}{l}{0.935 {\normalfont$\pm$} 0.005} & \multicolumn{2}{l}{0.930 {\normalfont$\pm$} 0.004} \\ \hline
Attention-WNET       &       & \multicolumn{2}{l}{0.977 {\normalfont$\pm$} 0.003} & \multicolumn{2}{l}{0.974 {\normalfont$\pm$} 0.006} \\ \hline
Swin-WNET            &       & \multicolumn{2}{l}{0.969 {\normalfont$\pm$} 0.009} & \multicolumn{2}{l}{0.969 {\normalfont$\pm$} 0.010} \\ \hline
\end{tabular}
\end{table}

\begin{table}[htb]
\caption{Results of Artery/ Vein Classification on DRIVE Dataset with Separate Metrics}
\label{tab:Binary_DRIVE_Dataset_new}
\begin{tabular}{llllll}
\hline
\textbf{Model}       &       & \multicolumn{2}{l}{\textbf{Artery Accuracy}} & \multicolumn{2}{l}{\textbf{Vein Accuracy}} \\ \hline
Attention-UNET       &       & \multicolumn{2}{l}{0.928 {\normalfont$\pm$} 0.007} & \multicolumn{2}{l}{0.931 {\normalfont$\pm$} 0.003} \\ \hline
Attention-WNET       &       & \multicolumn{2}{l}{0.970 {\normalfont$\pm$} 0.002} & \multicolumn{2}{l}{0.969 {\normalfont$\pm$} 0.003} \\ \hline
\end{tabular}
\end{table}

\subsection{Cross Dataset Results}

To evaluate the generalizability of our proposed \textbf{Attention-WNet} model, we conducted cross-dataset experiments for artery/vein classification using the DRIVE and HRF datasets. Specifically, we trained the model on the HRF dataset and tested on the DRIVE dataset, and vice versa, to assess its robustness across varying image domains and acquisition conditions.

Table~\ref{tab:Cross_Dataset_DRIVE} presents the results when the models were trained on the HRF dataset and evaluated on the DRIVE dataset. We report accuracy and F1 scores across three categories: (i) all vessel pixels, (ii) discovered vessel centerline pixels, and (iii) discovered centerline pixels of vessels wider than two pixels. The proposed Attention-WNet demonstrates competitive performance across all categories, closely matching or outperforming the baseline Swin-WNet, thereby showcasing its ability to generalize vessel structures effectively during cross-dataset evaluation.

Similarly, Table~\ref{tab:Cross_Dataset_HRF} shows the results when the models were trained on the DRIVE dataset and tested on the HRF dataset under the same evaluation conditions. The proposed Attention-WNet achieves higher accuracy and F1 scores compared to Swin-WNet across all evaluation categories in this setting, illustrating its robustness in transferring from a lower-resolution dataset (DRIVE) to a higher-resolution dataset (HRF).

Furthermore, Table~\ref{tab:Cross_Separate_Dataset} reports the per-class artery and vein classification accuracies under the cross-dataset settings. The proposed Attention-WNet achieves an artery accuracy of $0.968 \pm 0.004$ and a vein accuracy of $0.959 \pm 0.005$ when trained on HRF and tested on DRIVE, and achieves $0.966 \pm 0.001$ artery accuracy and $0.964 \pm 0.004$ vein accuracy when trained on DRIVE and tested on HRF. These results confirm the consistent per-class performance of the proposed model in unseen domains.

Overall, the cross-dataset results summarized in Tables~\ref{tab:Cross_Dataset_DRIVE}, \ref{tab:Cross_Dataset_HRF}, and \ref{tab:Cross_Separate_Dataset} demonstrate that the proposed Attention-WNet model generalizes effectively across datasets with different imaging conditions, providing reliable artery/vein classification performance and ensuring its suitability for real-world deployment in clinical settings.
\begin{table*}[htb]
\centering
\caption{Results of Artery/Vein Classification on Cross Dataset: Trained on HRF tested on DRIVE dataset}
\label{tab:Cross_Dataset_DRIVE}
\resizebox{\textwidth}{!}{%
\begin{tabular}{@{}l c cc cc cc@{}}
\toprule
Model &  & \multicolumn{2}{c}{\begin{tabular}[c]{@{}c@{}}All vessel\\ pixels\end{tabular}} 
      & \multicolumn{2}{c}{\begin{tabular}[c]{@{}c@{}}Discovered vessel\\ centerline pixels\end{tabular}} 
      & \multicolumn{2}{c}{\begin{tabular}[c]{@{}c@{}}Discovered centerline pixels of\\ vessels wider than two pixels\end{tabular}} \\
\cmidrule(lr){3-4} \cmidrule(lr){5-6} \cmidrule(lr){7-8}
      &  & Accuracy & F1 & Accuracy & F1 & Accuracy & F1 \\
\midrule
\textbf{Attention-WNet} & & \textbf{0.939 ± 0.004} & \textbf{0.925 ± 0.006} & \textbf{0.727 ± 0.058} & \textbf{0.716 ± 0.062} & \textbf{0.736 ± 0.061} & \textbf{0.724 ± 0.066} \\
Swin-WNet    & & 0.949 ± 0.006 & 0.940 ± 0.009 & 0.750 ± 0.032 & 0.746 ± 0.035 & 0.766 ± 0.031 & 0.762 ± 0.035 \\
\bottomrule
\end{tabular}%
}
\end{table*}

\vspace{1em} 

\begin{table*}[htb]
\centering
\caption{Results of Artery/Vein Classification on Cross Dataset: Trained on DRIVE tested on HRF dataset}
\label{tab:Cross_Dataset_HRF}
\resizebox{\textwidth}{!}{%
\begin{tabular}{@{}l c cc cc cc@{}}
\toprule
Model &  & \multicolumn{2}{c}{\begin{tabular}[c]{@{}c@{}}All vessel\\ pixels\end{tabular}} 
      & \multicolumn{2}{c}{\begin{tabular}[c]{@{}c@{}}Discovered vessel\\ centerline pixels\end{tabular}} 
      & \multicolumn{2}{c}{\begin{tabular}[c]{@{}c@{}}Discovered centerline pixels of\\ vessels wider than two pixels\end{tabular}} \\
\cmidrule(lr){3-4} \cmidrule(lr){5-6} \cmidrule(lr){7-8}
      &  & Accuracy & F1 & Accuracy & F1 & Accuracy & F1 \\
\midrule
\textbf{Attention-WNet}   & & \textbf{0.939 ± 0.005} & \textbf{0.940 ± 0.006} & \textbf{0.696 ± 0.035} & \textbf{0.686 ± 0.039} & \textbf{0.712 ± 0.033} & \textbf{0.707 ± 0.037} \\
Swin-WNet       & & 0.917 ± 0.022 & 0.925 ± 0.020 & 0.596 ± 0.013 & 0.593 ± 0.019 & 0.603 ± 0.014 & 0.594 ± 0.012 \\
\bottomrule
\end{tabular}%
}
\end{table*}

\begin{table}[htb]
\caption{Cross-Dataset Results of Artery/ Vein Classification between DRIVE and HRF}
\label{tab:Cross_Separate_Dataset}
\begin{tabular}{llllll}
\hline
\textbf{Model}       &       & \multicolumn{2}{l}{\textbf{Artery Accuracy}} & \multicolumn{2}{l}{\textbf{Vein Accuracy}} \\ \hline
\begin{tabular}[c]{@{}l@{}}Attention-WNET \\ (Trained on DRIVE, Tested on HRF)\end{tabular} & & \multicolumn{2}{l}{0.966 {\normalfont$\pm$} 0.001} & \multicolumn{2}{l}{0.964 {\normalfont$\pm$} 0.004} \\ \hline
\begin{tabular}[c]{@{}l@{}}Swin-WNET \\ (Trained on DRIVE, Tested on HRF)\end{tabular} & & \multicolumn{2}{l}{0.960 {\normalfont$\pm$} 0.003} & \multicolumn{2}{l}{0.958 {\normalfont$\pm$} 0.005} \\ \hline
\begin{tabular}[c]{@{}l@{}}Attention-WNET \\ (Trained on HRF, Tested on DRIVE)\end{tabular} & & \multicolumn{2}{l}{0.968 {\normalfont$\pm$} 0.004} & \multicolumn{2}{l}{0.959 {\normalfont$\pm$} 0.005} \\ \hline
\end{tabular}
\end{table}

\section{Conclusions}\label{sec5}
In conclusion, the proposed work introduced a novel approach that leverages the strengths of both the Attention U-Net and W-Net architectures to address the challenging task of artery and vein segmentation in medical images. By training separate models for arteries and veins and subsequently combining their outputs, a detailed artery-vein segmentation map has been obtained. The proposed strategy offers a promising alternative to the conventional segmentation methods.

The decision to train dedicated models for artery and vein segmentation comes from the understanding that arteries and veins exhibit unique structural characteristics and appearance patterns. The proposed approach capitalizes on the intrinsic dissimilarity, enabling the individual models to specialize in capturing the distinctive features of each vessel component. The subsequent concatenation of these specialized outputs results in a more accurate artery-vein segmentation, as opposed to a one-size-fits-all approach.

Furthermore, the inherent complexity of the artery-vein interaction within medical images presents challenges that are not easily overcome by a single architecture. The proposed hybrid approach addresses this complexity by utilizing the power of attention mechanisms and skip connections, allowing the models to effectively capture and emphasize critical features. While our architecture may not have surpassed all the current state-of-the-art models, it provides a significant advancement and gives valuable insights into the potential of dedicated artery and vein models.

As future research directions, the authors recommend exploring ways to refine the attention mechanisms and architectural components of our hybrid approach, potentially pushing its performance closer to the state-of-the-art. The proposed work thus contributes to the ongoing advancement of artery vein segmentation techniques and underscores the importance of different approaches in such complex segmentation tasks.

\section{Data Availability}
The DRIVE and HRF datasets used in the study are publicly available.\\\\
The DRIVE dataset can be downloaded from 

- https://drive.grand-challenge.org/ \\\\
The HRF dataset can be downloaded from

- https://www5.cs.fau.de/research/data/fundus-images/
\section{Declarations}

\subsection{Funding} The authors did not receive support from any organization for the submitted work.


\begin{thebibliography}{39}
\providecommand{\natexlab}[1]{#1}
\providecommand{\url}[1]{{#1}}
\providecommand{\urlprefix}{URL }
\providecommand{\doi}[1]{\url{https://doi.org/#1}}
\providecommand{\eprint}[2][]{\url{#2}}
 \bibcommenthead

\bibitem[{AlBadawi and Fraz(2018)}]{albadawi2018arterioles}
AlBadawi S, Fraz M (2018) Arterioles and venules classification in retinal images using fully convolutional deep neural network. In: International conference image analysis and recognition, Springer, pp 659--668

\bibitem[{Badrinarayanan et~al(2017)Badrinarayanan, Kendall, and Cipolla}]{badrinarayanan2017segnet}
Badrinarayanan V, Kendall A, Cipolla R (2017) Segnet: A deep convolutional encoder-decoder architecture for image segmentation. IEEE Transactions on Pattern Analysis and Machine Intelligence 39(12):2481--2495

\bibitem[{Chowdhury et~al(2022)Chowdhury, Mann, Morgan, Vukmirovic, Mehnert, and Sohel}]{Chowdhury2022}
Chowdhury AE, Mann G, Morgan WH, et~al (2022) Msganet-rav: A multiscale guided attention network for artery-vein segmentation and classification from optic disc and retinal images. Journal of Optometry 15:S58--S69

\bibitem[{Dashtbozorg et~al(2013)Dashtbozorg, Mendon{\c{c}}a, and Campilho}]{dashtbozorg2013automatic}
Dashtbozorg B, Mendon{\c{c}}a AM, Campilho A (2013) An automatic graph-based approach for artery/vein classification in retinal images. IEEE Transactions on Image Processing 23(3):1073--1083

\bibitem[{De et~al(2017)De, Zhang, Lin, and Cheng}]{de2017transduction}
De J, Zhang X, Lin F, et~al (2017) Transduction on directed graphs via absorbing random walks. IEEE transactions on pattern analysis and machine intelligence 40(7):1770--1784

\bibitem[{Estrada et~al(2014)Estrada, Tomasi, Schmidler, and Farsiu}]{estrada2014tree}
Estrada R, Tomasi C, Schmidler SC, et~al (2014) Tree topology estimation. IEEE transactions on pattern analysis and machine intelligence 37(8):1688--1701

\bibitem[{Estrada et~al(2015)Estrada, Allingham, Mettu, Cousins, Tomasi, and Farsiu}]{estrada2015retinal}
Estrada R, Allingham MJ, Mettu PS, et~al (2015) Retinal artery-vein classification via topology estimation. IEEE transactions on medical imaging 34(12):2518--2534

\bibitem[{Galdran et~al(2019)Galdran, Meyer, Costa, Campilho et~al}]{galdran2019uncertainty}
Galdran A, Meyer M, Costa P, et~al (2019) Uncertainty-aware artery/vein classification on retinal images. In: 2019 IEEE 16th International Symposium on Biomedical Imaging (ISBI 2019), IEEE, pp 556--560

\bibitem[{Galdran et~al(2020)Galdran, Anjos, Dolz, Chakor, Lombaert, and Ayed}]{galdran2020little}
Galdran A, Anjos A, Dolz J, et~al (2020) The little w-net that could: state-of-the-art retinal vessel segmentation with minimalistic models. arXiv preprint arXiv:200901907

\bibitem[{Hemelings et~al(2019)Hemelings, Elen, Stalmans, Van~Keer, De~Boever, and Blaschko}]{hemelings2019artery}
Hemelings R, Elen B, Stalmans I, et~al (2019) Artery--vein segmentation in fundus images using a fully convolutional network. Computerized Medical Imaging and Graphics 76:101636

\bibitem[{Hu et~al(2021{\natexlab{a}})Hu, Wang, Cao, Wu, Jonas, Wang, and Zhang}]{hu2021automatic}
Hu J, Wang H, Cao Z, et~al (2021{\natexlab{a}}) Automatic artery/vein classification using a vessel-constraint network for multicenter fundus images. Frontiers in cell and developmental biology p 1194

\bibitem[{Hu et~al(2021{\natexlab{b}})Hu, Wang, Cao, Wu, Jonas, Wang, and Zhang}]{Hu2021}
Hu J, Wang H, Cao Z, et~al (2021{\natexlab{b}}) Automatic artery/vein classification using a vessel-constraint network for multicenter fundus images. Frontiers in Cell and Developmental Biology 9:659941

\bibitem[{Hu et~al(2022)Hu, Wang, Wu, Cao, Mou, Zhao, and Zhang}]{hu2022multi}
Hu J, Wang H, Wu G, et~al (2022) Multi-scale interactive network with artery/vein discriminator for retinal vessel classification. IEEE Journal of Biomedical and Health Informatics

\bibitem[{Joshi et~al(2011)Joshi, Garvin, Reinhardt, and Abramoff}]{joshi2011automated}
Joshi VS, Garvin MK, Reinhardt JM, et~al (2011) Automated method for the identification and analysis of vascular tree structures in retinal vessel network. In: Medical Imaging 2011: Computer-Aided Diagnosis, SPIE, pp 143--153

\bibitem[{Joshi et~al(2014)Joshi, Reinhardt, Garvin, and Abramoff}]{joshi2014automated}
Joshi VS, Reinhardt JM, Garvin MK, et~al (2014) Automated method for identification and artery-venous classification of vessel trees in retinal vessel networks. PloS one 9(2):e88061

\bibitem[{Kirbas and Quek(2004)}]{kirbas2004review}
Kirbas C, Quek F (2004) A review of vessel extraction techniques and algorithms. ACM Computing Surveys (CSUR) 36(2):81--121

\bibitem[{Lepetit-Aimon et~al(2018)Lepetit-Aimon, Duval, and Cheriet}]{lepetit2018large}
Lepetit-Aimon G, Duval R, Cheriet F (2018) Large receptive field fully convolutional network for semantic segmentation of retinal vasculature in fundus images. In: Computational Pathology and Ophthalmic Medical Image Analysis. Springer, p 201--209

\bibitem[{Li et~al(2019)Li, Liu, Chen, Jiang, Zhang, and Wu}]{li2019automatic}
Li F, Liu Z, Chen H, et~al (2019) Automatic detection of diabetic retinopathy in retinal fundus photographs based on deep learning algorithm. Translational vision science \& technology 8(6):4--4

\bibitem[{Li et~al(2020)Li, Verma, Nakashima, Kawasaki, and Nagahara}]{li2020joint}
Li L, Verma M, Nakashima Y, et~al (2020) Joint learning of vessel segmentation and artery/vein classification with post-processing. In: Medical Imaging with Deep Learning, PMLR, pp 440--453

\bibitem[{Lin et~al(2017)Lin, Goyal, Girshick, He, and Doll{\'a}r}]{lin2017focal}
Lin TY, Goyal P, Girshick R, et~al (2017) Focal loss for dense object detection. In: Proceedings of the IEEE international conference on computer vision, pp 2980--2988

\bibitem[{Martinez-Perez et~al(2002)Martinez-Perez, Highes, Stanton, Thorn, Chapman, Bharath, and Parker}]{martinez2002retinal}
Martinez-Perez ME, Highes A, Stanton AV, et~al (2002) Retinal vascular tree morphology: a semi-automatic quantification. IEEE Transactions on Biomedical Engineering 49(8):912--917

\bibitem[{Mitta et~al(2020)Mitta, Chatterjee, Speck, and N{\"u}rnberger}]{mitta2020upgraded}
Mitta D, Chatterjee S, Speck O, et~al (2020) Upgraded w-net with attention gates and its application in unsupervised 3d liver segmentation. arXiv preprint arXiv:201110654

\bibitem[{Oktay et~al(2018)Oktay, Schlemper, Folgoc, Lee, Heinrich, Misawa, Mori, McDonagh, Hammerla, Kainz et~al}]{oktay2018attention}
Oktay O, Schlemper J, Folgoc LL, et~al (2018) Attention u-net: Learning where to look for the pancreas. arXiv preprint arXiv:180403999

\bibitem[{Ooi et~al(2021)Ooi, Embong, Abd~Hamid, Zainon, Wang, Ng, Hamzah, Teoh, and Ibrahim}]{ooi2021interactive}
Ooi AZH, Embong Z, Abd~Hamid AI, et~al (2021) Interactive blood vessel segmentation from retinal fundus image based on canny edge detector. Sensors 21(19):6380

\bibitem[{Organization et~al(2019)}]{world2019world}
Organization WH, et~al (2019) World report on vision

\bibitem[{Orlando et~al(2020)Orlando, Fu, Breda, van Keer, Bathula, Diaz-Pinto, Fang, Heng, Kim, Lee et~al}]{orlando2020refuge}
Orlando JI, Fu H, Breda JB, et~al (2020) Refuge challenge: A unified framework for evaluating automated methods for glaucoma assessment from fundus photographs. Medical image analysis 59:101570

\bibitem[{Ronneberger et~al(2015{\natexlab{a}})Ronneberger, Fischer, and Brox}]{ronneberger2015unet}
Ronneberger O, Fischer P, Brox T (2015{\natexlab{a}}) U-net: Convolutional networks for biomedical image segmentation. In: Medical Image Computing and Computer-Assisted Intervention–MICCAI 2015: 18th International Conference, Munich, Germany, October 5-9, 2015, Proceedings, Part III. Springer International Publishing, pp 234--241

\bibitem[{Ronneberger et~al(2015{\natexlab{b}})Ronneberger, Fischer, and Brox}]{ronneberger2015u}
Ronneberger O, Fischer P, Brox T (2015{\natexlab{b}}) U-net: Convolutional networks for biomedical image segmentation. In: International Conference on Medical image computing and computer-assisted intervention, Springer, pp 234--241

\bibitem[{Rothaus et~al(2009)Rothaus, Jiang, and Rhiem}]{rothaus2009separation}
Rothaus K, Jiang X, Rhiem P (2009) Separation of the retinal vascular graph in arteries and veins based upon structural knowledge. Image and Vision Computing 27(7):864--875

\bibitem[{Saleh et~al(2011)Saleh, Eswaran, and Mueen}]{saleh2011automated}
Saleh MD, Eswaran C, Mueen A (2011) An automated blood vessel segmentation algorithm using histogram equalization and automatic threshold selection. Journal of digital imaging 24:564--572

\bibitem[{Shen et~al(2018)Shen, Li, Xie, and Zhao}]{shen2018retinal}
Shen H, Li X, Xie X, et~al (2018) Retinal blood vessel segmentation and classification using deep convolutional neural network. In: 2018 7th International Conference on Computer Science and Engineering (ICCSE), IEEE, pp 430--435

\bibitem[{Soares et~al(2006)Soares, Leandro, Cesar, Jelinek, and Cree}]{soares2006retinal}
Soares JV, Leandro JJ, Cesar RM, et~al (2006) Retinal vessel segmentation using the 2-d gabor wavelet and supervised classification. IEEE Transactions on medical Imaging 25(9):1214--1222

\bibitem[{Staal et~al(2004)Staal, Abr{\`a}moff, Niemeijer, Viergever, and van Ginneken}]{staal2004automated}
Staal J, Abr{\`a}moff MD, Niemeijer M, et~al (2004) Automated segmentation of retinal blood vessels in fundus images. IEEE transactions on medical imaging 23(4):501--509

\bibitem[{Vashist et~al(2022)Vashist, Senjam, Gupta, Gupta, Shamanna, Wadhwani, Shukla, Manna, Yadav, and Bharadwaj}]{vashist2022blindness}
Vashist P, Senjam SS, Gupta V, et~al (2022) Blindness and visual impairment and their causes in india: Results of a nationally representative survey. Plos one 17(7):e0271736

\bibitem[{Wang et~al(2020)Wang, Jiang, Liu, Cheng, and Yang}]{wang2020multi}
Wang Z, Jiang X, Liu J, et~al (2020) Multi-task siamese network for retinal artery/vein separation via deep convolution along vessel. IEEE Transactions on Medical Imaging 39(9):2904--2919

\bibitem[{Welikala et~al(2017)Welikala, Foster, Whincup, Rudnicka, Owen, Strachan, Barman et~al}]{welikala2017automated}
Welikala R, Foster P, Whincup P, et~al (2017) Automated arteriole and venule classification using deep learning for retinal images from the uk biobank cohort. Computers in biology and medicine 90:23--32

\bibitem[{Xia and Kulis(2017)}]{xia2017wnet}
Xia X, Kulis B (2017) W-net: A deep model for fully unsupervised image segmentation. arXiv preprint arXiv:171108506

\bibitem[{Yin et~al(2019)Yin, Irshad, and Zhang}]{yin2019artery}
Yin XX, Irshad S, Zhang Y (2019) Artery/vein classification of retinal vessels using classifiers fusion. Health Information Science and Systems 7(1):1--14

\bibitem[{Zhao et~al(2019)Zhao, Xie, Zhang, Zheng, Zhao, Qi, Zhao, Su, Liu, and Liu}]{zhao2019retinal}
Zhao Y, Xie J, Zhang H, et~al (2019) Retinal vascular network topology reconstruction and artery/vein classification via dominant set clustering. IEEE transactions on medical imaging 39(2):341--356

\end{thebibliography}
\end{document}